\documentclass[10pt,twocolumn]{article}

\usepackage[square,numbers]{natbib}
\usepackage{cvpr}
\usepackage{xcolor,xspace}
\usepackage{algorithm}
\usepackage[noend]{algpseudocode}
\usepackage{amsmath}
\usepackage{amssymb}
\usepackage{booktabs}
\usepackage{balance}
\usepackage{float}
\usepackage{amsfonts}      
\usepackage{url}

\newcommand{\comment}[1]{}

\usepackage{times}
\usepackage{epsfig}
\usepackage{graphicx}

\usepackage{subfigure}

\usepackage[pagebackref=true,breaklinks=true,colorlinks,bookmarks=false]{hyperref}

\cvprfinalcopy 

\def\cvprPaperID{XXXXX} 

\ifcvprfinal\fi
\begin{document}

\title{Accelerated AI Inference via Dynamic Execution Methods}
\author{Haim Barad\\
{\tt\small haim.barad@intel.com}
\and
Jascha Achterberg\\
{\tt\small jascha.achterberg@dpag.ox.ac.uk}
\and
Tien Pei Chou\\
{\tt\small joey.t.p.chou@gmail.com}
\and
Jean Yu\\
{\tt\small jean1.yu@intel.com}
}

\maketitle
\thispagestyle{empty}

\begin{abstract}

In this paper, we focus on Dynamic Execution techniques that optimize the computation flow based on input. This aims to identify simpler problems that can be solved using fewer resources, similar to human cognition. The techniques discussed include early exit from deep networks, speculative sampling for language models, and adaptive steps for diffusion models. Experimental results demonstrate that these dynamic approaches can significantly improve latency and throughput without compromising quality. When combined with model-based optimizations, such as quantization, dynamic execution provides a powerful multi-pronged strategy to optimize AI inference.

Generative AI requires a large amount of compute resources. This is expected to grow, and demand for resources in data centers through to the edge is expected to continue to increase at high rates. We take advantage of existing research and provide additional innovations for some generative optimizations. In the case of LLMs, we provide more efficient sampling methods that depend on the complexity of the data. In the case of diffusion model generation, we provide a new method that also leverages the difficulty of the input prompt to predict an optimal early stopping point.

Therefore, dynamic execution methods are relevant because they add another dimension of performance optimizations. Performance is critical from a competitive point of view, but increasing capacity can result in significant power savings and cost savings. We have provided several integrations of these techniques into several Intel performance libraries and Huggingface Optimum. These integrations will make them easier to use and increase the adoption of these techniques.

\end{abstract}


\section{Introduction}

The computational requirements for AI inference are growing at a rapid rate. Larger models and increased demand (i.e. queries directly or via APIs) in these models drive higher inference costs \cite{kaplan_scaling_2020}\cite{schwartz_green_2019}. To control the energy demands and hardware requirements of the newest generation AI models, we must find ways to run large-scale AI architectures more efficiently while achieving the same accuracy and performance levels. Past years of research have resulted in new methods to both make models more efficient by accelerating the inference of models and making the models themselves more lightweight. 

When thinking about the efficiency of models, we want to optimize the computational resources needed to achieve a given output of a computation. Given a specific output of a computational / inference process, we can aim to tweak the power consumption, throughput, and latency of a target architecture. In this paper, we are starting with an overview of very established model compression techniques (quantization, sparsity), which can address power consumption through lowering the memory demand. However, our primary focus is on methods to improve model throughput by allowing models to take shortcuts during the inference process. Note that model throughput is a very interesting target, as it is directly correlated with the total costs of serving the model as a service. Alongside a review of established techniques, we will also provide an overview of novel results on early exit, speculative sampling in token and feature space techniques in large language models (LLMs) and techniques to accelerate diffusion models. We show that significant throughput improvements can be achieved by expanding established model architectures with the methods presented here. Out of the scope of this paper are that many architectural improvements can be made given the representation (e.g. ONNX) and runtime engines.

We believe that significant improvements that can be achieved through the inference optimization techniques presented here should not be perceived as the result of arbitrary tricks of model optimization. Instead, they all have deep roots in the study of intelligent systems. John von Neumann has already discussed that we should be able to build intelligent systems with low numerical precision given how noisy signal communication is in the brain \cite{von_neumann_computer_1958}. Similarly, large and powerful brains of mammals also achieve energy efficiency through sparsification of their brain networks \cite{bullmore_economy_2012}\cite{achterberg_spatially_2023}. Lastly, the idea of dynamic execution is deeply rooted in cognitive science, where humans and animals have been observed to choose between different problem solving strategies considering the mental effort and computational resources needed to execute a given strategy \cite{lewis_computational_2014}\cite{gershman_computational_2015}\cite{kool_mental_2018}. Given that a computationally cheaper strategy achieves a comparable result, they would quickly change course to pursue the simpler strategy. The inference optimization techniques presented in the following hence do not only show promising efficiency effects but also show how there is a joint mechanism of efficient information processing connecting artificial and biological computing systems.

\section{Model Compression: Approximate Math}

Let’s dive into the concepts of \textbf{quantization} and \textbf{sparsity} in the context of model compression. By no means are these the only methods to compress a model's size, but we will discuss these in the context of comparing model-based optimizations and dynamic execution.

\subsection{Quantization}

An excellent overview of both quantization and pruning techniques has been published by Liang et al. \cite{Liang2021}. These powerful techniques provide for improved performance by essentially compressing the model: by quantizing parameters to a lower number of bits and/or eliminating redundant computations.
Both of these methods improve the overall compute requirements at inference time, since the model is compressed. This compression can be achieved by a variety of methods, but they all come down to a brute-force method to approximate the uncompressed computation so that the accuracy loss is minimized.
In other words, I classify these categories as `approximate math' - we compress and increase the aggressiveness of compression techniques as long as accuracy degradation is still tolerable. 

If you think about it in a generic way, many neural network topologies compute a function:
$$y = F(x)$$

which already is some approximation that might help accomplish a task. For example, if this is a classification problem, the function F(x) might represent a decision boundary between classes, which is computed as a composite of layers. So, while F(x) might already be, in some sense, an approximation, we consider it to be the most accurate representation of the model.
With some approaches that fall into the approximate-math category, we instead compute:

$$y' = F'(x)$$

where $F'$ is some approximation of the full computational version of the model $F(x)$. Therefore, the goal of these approximate math techniques is to minimize the difference between the full math model and the model that has been compressed.

$$L_c = min |y'-y|$$

This would represent a measure of loss in the compression method. Often, we find these compression methods to be valuable, as the compression loss $L_c$ is often tolerable and well worth the benefits of a smaller and more efficient model.

Note that there are some additional related optimizations (e.g. factorization), but these still fit within a brute-force approximate math approach.
While the rest of this article will focus on dynamic execution optimizations (notably Early Exit), we should recognize that these two categories of optimizations can be used together to provide a completely optimized solution in many respects (latency, throughput, power consumption, cost, etc.).

\subsection{Sparsity}

Sparsity, on the other hand, is a technique that aims to make most weights of the neural network zero. Based on observation, in many models, only a small part of the weights contribute significantly to the final predictions.

Mathematically, sparsity can be induced by adding a regularization term to the loss function during training. This term penalizes the model for having large weights, effectively pushing many of them towards zero.

Once the model is sparse, it can be compressed by simply storing the locations and values of the nonzero weights. This significantly reduces the memory footprint of the model, as the zero weights do not need to be stored.

Like quantization, sparsity introduces some approximation error. However, if the original model has redundant weights (which is often the case), this error can be minimal. However, it should be noted that this approximation error can be limited to the error in the training data, while the validation error might decrease due to a better generalization performance of the model \cite{hardt_patterns_2021}.

In conclusion, both quantization and sparsity are powerful techniques for compressing neural networks. They work by introducing mathematical approximations that reduce the memory footprint and computational requirements of the model, often with minimal impact on accuracy. These techniques are particularly useful for deploying models on resource-constrained devices such as mobile phones and embedded systems.

\subsection{Results on Model Compression}
It is beyond the scope of this paper to dive into model compression techniques. Substantial research and tools are widely available to exploit techniques such as quantization and pruning \cite{Liang2021}. Often compressed models are publicly available on hubs, such as HuggingFace, so that further compression is often unnecessary.

Since these benefits of the techniques often come from the compression itself, there is an inherent speedup just from the compressed nature of the optimized model. These methods are covered elsewhere and beyond the scope of this paper.

\section{Dynamic Execution: Taking Short Cuts}

In the dynamic execution scenario, we employ a set of specialized techniques designed to scrutinize the specific query at hand. These techniques involve a thorough examination of the query's structure, content, and context with the aim of discerning whether the problem it represents can be addressed in a more straightforward manner.

This approach mirrors the way humans tackle problem-solving. Just as we, as humans, are often able to identify problems that are 'easy' or 'simple' and solve them with less effort compared to 'hard' or 'complex' problems, these techniques strive to do the same. They are designed to recognize simpler problems and solve them more efficiently, thereby saving computational resources and time.

We refer to this approach as Dynamic Execution. The term 'dynamic' signifies the adaptability and flexibility of this approach. Unlike static methods that rigidly adhere to a predetermined path regardless of the problem's nature, Dynamic Execution adjusts its strategy based on the specific problem it encounters, that is, the opportunity is data dependent.

The goal of Dynamic Execution is not to optimize the model itself, but to optimize the compute flow. In other words, it seeks to streamline the process through which the model interacts with the data. By tailoring the compute flow to the data presented to the model, Dynamic Execution ensures that the model's computational resources are utilized in the most efficient manner possible.

In essence, Dynamic Execution is about making the problem-solving process as efficient and effective as possible by adapting the strategy to the problem at hand, much like how humans approach problem-solving. It is about working smarter, not harder. This approach not only saves computational resources but also improves the speed and accuracy of the problem-solving process.

\subsection{Dynamic Execution Explained}

Early Exit is a strategy with a simple and easy-to-understand concept, and we will use this to explain the methodology of dynamic execution. We then explore Early Exit and other techniques aimed at Generative AI and discuss the results.

\begin{figure*}
  \centering
 \includegraphics[width=1\linewidth]{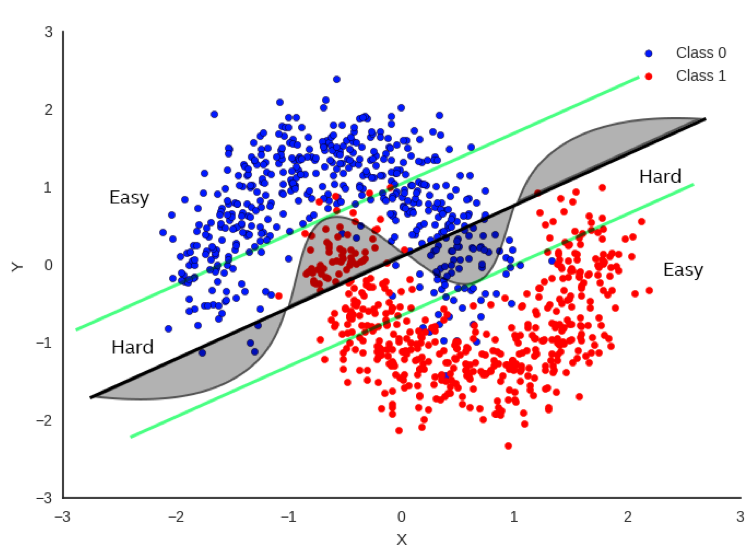}
  \caption{Simple and more expressive classification boundaries}
  \label{fig:HardEasyClassification}
\end{figure*}

Figure \ref{fig:HardEasyClassification} shows a simple example in a 2-D feature space. Although a deep network can represent a more complex and expressive boundary between classes (as shown in the curved area), it is also clear that much of the data can be properly classified with even the simplest of classification boundaries (e.g., even a linear classifier as shown by the straight black line). In other words, data points outside the two parallel green lines that bound the complex decision boundary of the deep network can be accurately classified even with a simple linear decision boundary. Points within the two parallel green lines are more difficult to distinguish and require extra processing to accurately classify. We use a confidence measure to determine whether we have enough information to confidently classify the data point at that stage.

Early Exit is a category of techniques based on this simple principle of doing 'enough work' given the difficulty of the input. There are other related methods that we will look at in this paper, including methods for generative sampling and early stopping of diffusion models.

In many respects, the 'Short-Cut Thinking' approach closely matches human intuition, which often seeks shortcuts to problem solving, separately treating 'easy' problems in a different manner from 'hard' problems. This coarse classification of the input data between easy and hard is done to quickly determine when and how shortcuts can be made to the computation, knowing that easy problems can often be treated with significantly less work.

Keep in mind that with Dynamic Execution, we do not change the model. We change how that model is computed by often doing fewer calculations (perhaps skipping unnecessary computations).

Note that the two optimization paradigms of model compression and dynamic execution are not mutually exclusive. There is no reason why we cannot apply multiple methodologies simultaneously to provide a sound optimization approach to the inference task.

In this paper, we will focus on dynamic execution methods for inference optimizations, with a focus on Generative AI. Some example methods include:

\begin{itemize}
    \item Early Exit - This technique involves adding exits at various stages in a deep neural network (DNN). The idea is to allow the network to terminate the inference process earlier for simpler tasks, thus saving computational resources. It takes advantage of the observation that some test examples can be easier to predict than others \cite{liao_global_2021}, \cite{ilhan_adaptive_2023}.

    \item Speculative Sampling - This method aims to speed up the inference process by computing several candidate tokens from a smaller draft model. These candidate tokens are then evaluated in parallel in the full target model \cite{leviathan_fast_2023}, \cite{barad_leveraging_2023}.

    \item Lookahead Decoding - This is a parallel decoding algorithm designed to accelerate Large Language Model (LLM) inference. The method breaks the sequential dependency in autoregressive decoding by simultaneously extracting and verifying n-grams (cached) directly with the LLM \cite{fu_break_2024}, \cite{noauthor_break_nodate}.

    \item Medusa - Medusa is a method that augments LLM inference by adding extra decoding heads to predict multiple subsequent tokens in parallel. These heads each produce multiple likely words for the corresponding position. These options are then combined and processed using a tree-based attention mechanism \cite{cai_medusa_2024}.

    \item EAGLE - (Extrapolation Algorithm for Greater Language-model Efficiency): EAGLE is a new baseline for fast decoding of LLMs with provable performance maintenance. This approach involves autoregression at the feature level (instead of the token level) of LLM, allowing a significant increase in generation efficiency \cite{li_eagle_2024}. An improved version of Eagle dynamically adjusts the tree structure of the candidate tokens using confidence levels, for even better performance \cite{li_eagle-2_2024}.

    \item Contextual Sparsity - This technique exploits the existence of contextual sparsity, which are small, input-dependent sets of attention heads and MLP parameters that yield approximately the same output as the dense model for a given input. It can be accurately predicted and exploited to speed up LLM inference in wall-clock time without compromising LLM's quality or in-context learning ability \cite{liu_deja_2023}.

    \item StepSaver: Early Stopping for Diffusion Generation, using an innovative NLP model specifically fine-tuned to determine the minimal number of denoising steps required for any given text prompt. This advanced model serves as a real-time tool that recommends the ideal denoise steps for generating high-quality images efficiently. It is designed to work seamlessly with the Diffusion model, ensuring that images are produced with superior quality in the shortest possible time. \cite{yu_step_2024}

    \item LLM Routing: Routing is a form of classification in which the prompt is used to determine the best model. The prompt is then routed to this model. By \textit{best}, we can use different criteria to determine the most effective model in terms of cost and accuracy. In many ways, routing is a form of dynamic execution done at the pipeline level where many of the other optimizations we are focusing on in this paper is done to make each LLM more efficient. For example, RouteLLM is an open-source framework for serving LLM routers and provides several mechanisms for reference, such as matrix factorization. \cite{ong_routellm_2024}
\end{itemize}

Please note that these are high-level descriptions, and the actual implementation of these methods can be complex and may vary depending on the specific use case and model architecture.

\subsection{Experiments}
This section will briefly show the results of experiments from several different types of dynamic execution. These results were reproduced on Intel platforms. It is beyond the scope of this paper to go into full details of each of the techniques, and references for each of these will be given should the reader want further details.

\subsection{Early Exit Results}
\begin{figure*}
  \centering
  \includegraphics[width=1\linewidth]{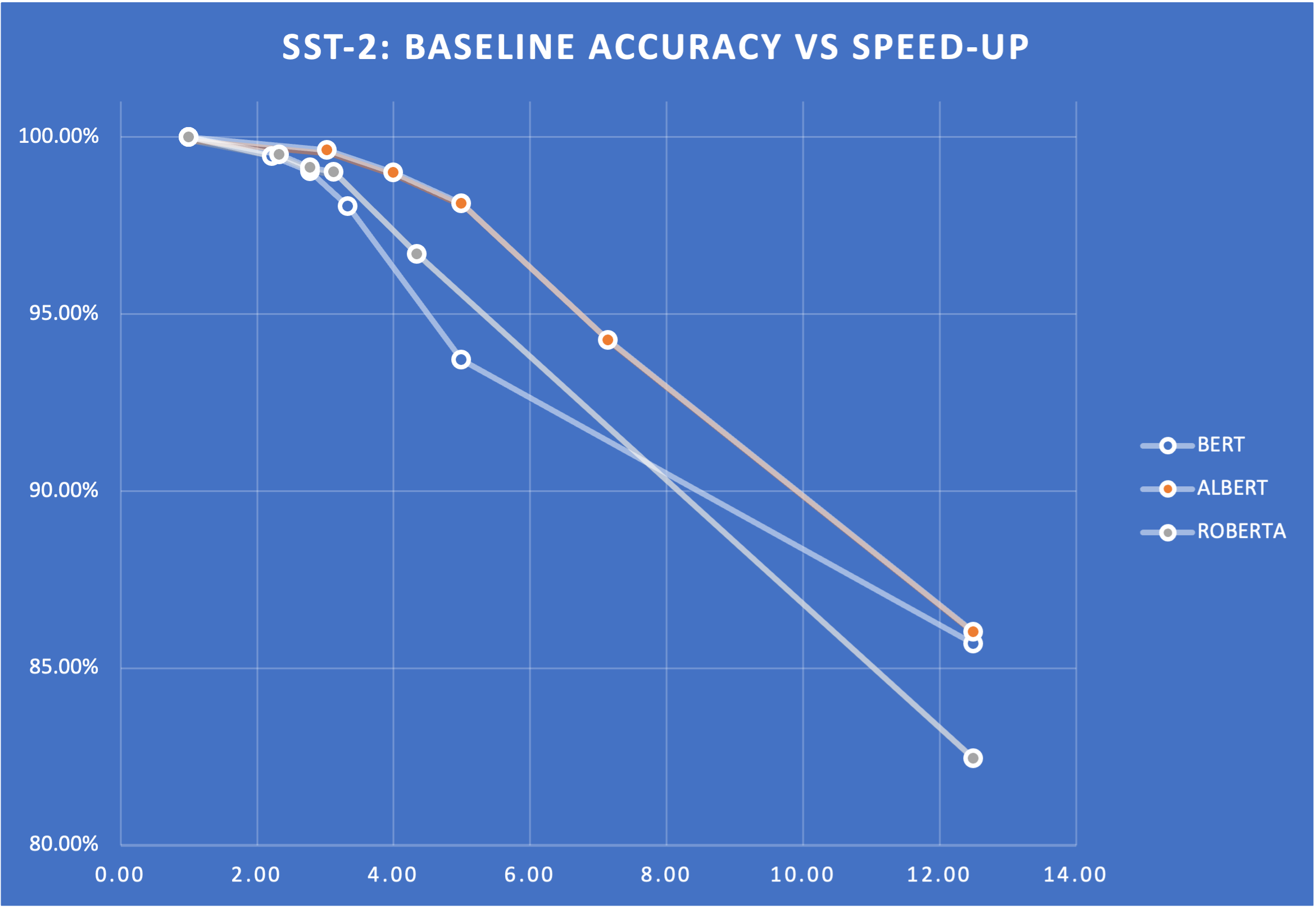}
  \caption{Early Exit: SST-2}
\label{fig:early_exit_scores}
\end{figure*}

We made use of the Early Exit strategy in several encoder models, including BERT, ROBERTA, and ALBERT. We measured the speed-ups on glue scores for various entropy thresholds. Figure \ref{fig:early_exit_scores} shows a plot of these scores and how they drop with respect to the entropy threshold. The scores show the percentage of the baseline score (that is, without Early Exit). Note that we can get 2x to 4X speed-up without sacrificing much quality.

\subsection{Speculative Sampling Results}
Speculative sampling is a technique designed to accelerate the decoding process of large language models \cite{chen_accelerating_2023}, \cite{mody_speculative_2023}. The concept behind speculative sampling is based on the observation that the latency of parallel scoring of short continuations, generated by a faster but less powerful draft model, is comparable to that of sampling a single token from the larger target model. This approach allows multiple tokens to be generated from each transformer call, increasing the speed of the decoding process.

The process of speculative sampling involves two models: a smaller, faster draft model and a larger, slower target model. The draft model speculates what the output is several steps into the future, while the target model determines how many of those tokens we should accept. The draft model decodes several tokens in a regular autoregressive fashion, and the probability outputs of the target and the draft models on the new predicted sequence are compared. Based on some rejection criteria, it is determined how many of the speculated tokens we want to keep. If a token is rejected, it is resampled using a combination of the two distributions, and no more tokens are accepted. If all speculated tokens are accepted, an additional final token can be sampled from the target model probability output.

In terms of performance boost, speculative sampling has shown significant improvements. For instance, it was benchmarked with Chinchilla, a 70 billion parameter language model, achieving a 2-2.5x decoding speedup in a distributed setup, without compromising the sample quality or making modifications to the model itself. Another example is the application of speculative decoding to Whisper, a general purpose speech transcription model, which resulted in a 2x speed-up in inference throughput \cite{gante_joao_assisted_2023}, \cite{gandhi_speculative_nodate}. Note that speculative sampling can be used to boost CPU inference performance, but the boost will likely be less (typically around 1.5x).

In conclusion, speculative sampling is a promising technique that leverages the strengths of both a draft and a target model to accelerate the decoding process of large language models. It offers a significant performance boost, making it a valuable tool in the field of natural language processing. However, it is important to note that the actual performance boost can vary depending on the specific models and setup used.

\subsection{EAGLE}

EAGLE, which stands for Extrapolation Algorithm for Greater Language-model Efficiency, is a novel form of speculative sampling designed to accelerate the decoding process of Large Language Models (LLMs) \cite{li_eagle_2024}. The key principle behind EAGLE is that the sequence of LLM feature vectors is compressible over time, making the prediction of subsequent feature vectors from previous ones easier¹.

Unlike traditional speculative sampling methods, EAGLE operates the drafting process auto-regressively at the second-to-top-layer feature level. This approach is more straightforward than working at the token level. EAGLE addresses the inherent uncertainty in feature-level autoregression by incorporating a token sequence advanced by one time step. This effectively resolves the uncertainty, enabling precise second-to-top-layer feature prediction with minimal overhead \cite{li_eagle_2024}.

\begin{figure*}
  \centering
  \includegraphics[width=1\linewidth]{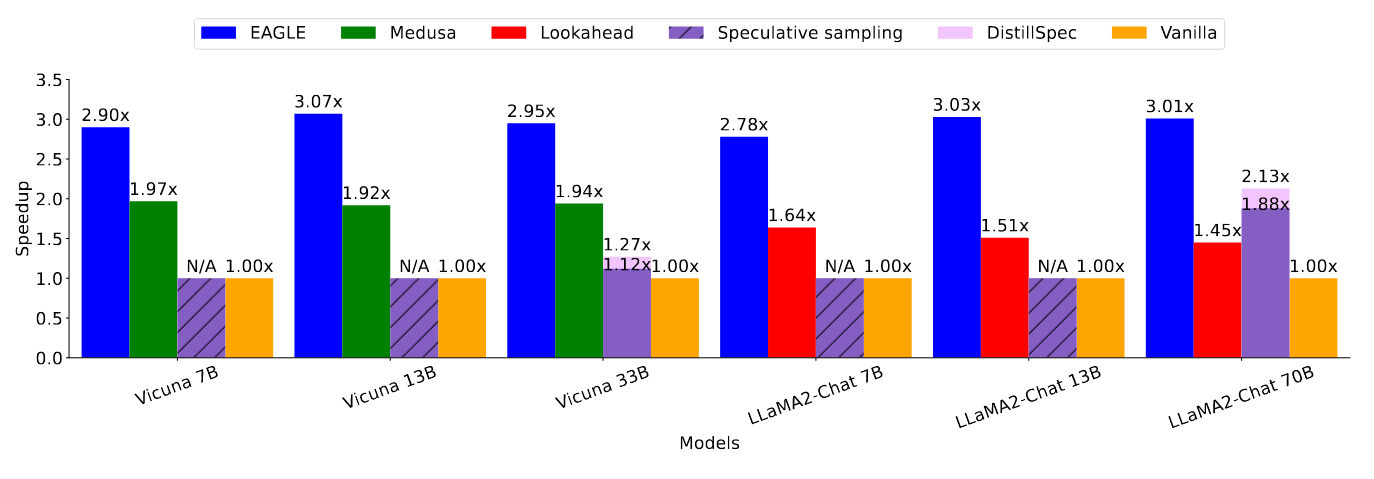}
  \caption{EAGLE Performance}
\label{fig:eagle_performance}
\end{figure*}

In terms of a performance boost (increase in throughput and lower latency), EAGLE has shown significant improvements. It achieves a 2x speedup on gpt-fast, one of the fastest-known open-source inference libraries. It is 3x faster than autoregressive sampling (13B), 2x faster than Lookahead (13B), and 1.6x faster than Medusa (13B). See figure \ref{fig:eagle_performance}. Moreover, EAGLE provably maintains consistency with vanilla decoding in the distribution of generated texts. For LLaMA2-Chat 70B, EAGLE achieved a latency speedup ratio of 2.7x-3.5x, doubled throughput, while maintaining the distribution of generated text \cite{li_eagle_2024}.

\textbf{[@Joey - please add a short paragraph or two regarding your results on PVC]}

In conclusion, EAGLE is a promising speculative sampling technique that provides a significant boost in inference performance. It maintains the original text distribution while accelerating the generation process. However, the actual performance boost can vary depending on the specific models and setup used.

\subsection{StepSaver for Diffusion Models}

Diffusion models iteratively enhance a random noise signal until it closely resembles the target data distribution \cite{notomoro_diffusion_2024}. When generating visual content such as images or videos, diffusion models have demonstrated significant realism \cite{hoppe_diffusion_2022}. For example, video diffusion models and SinFusion represent instances of diffusion models utilized in video synthesis \cite{ho_video_2022}\cite{nikankin_sinfusion_2023}. More recently, there has been growing attention towards models like OpenAI's Sora; however, this model is currently not publicly available due to its proprietary nature.

Performance in diffusion models involves a large number of iterations to recover images or videos from Gaussian noise \cite{chen_hierarchical_2023}. This process is called denoising and is trained on a specific number of iterations of denoising. The number of iterations in this sampling procedure is a key factor in the quality of the generated data, as measured by metrics, such as FID.

Latent space diffusion inference uses iterations in feature space, and performance suffers from the expense of many iterations required for quality output. Various techniques, such as patching transformation and transformer-based diffusion models \cite{peebles_scalable_2023}, improve the efficiency of each iteration.

\begin{figure*}
  \centering
  \includegraphics[width=1\linewidth]{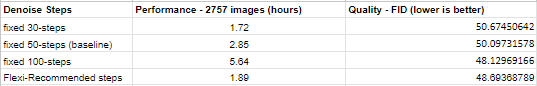}
  \caption{StepSaver Performance}
\label{fig:stepsaver_performance}
\end{figure*}

StepSaver dynamically recommends significantly lower denoising steps, which is critical to address the slow sampling issue of stable diffusion models during image generation \cite{yu_step_2024}. The recommended steps also ensure better image quality. Figure \ref{fig:stepsaver_performance} shows that images generated using dynamic steps result in a 3X throughput improvement and a similar image quality compared to static 100 steps. 

\section{Discussion}
In this paper, we discuss the methodology of dynamic execution methods in the context of inference optimizations. Together with model-based optimizations, dynamic execution offers opportunities that are data-dependent and do not conflict with model-based methods. 

We discussed several methods to optimize inference performance for different tasks, including Generative AI that are successful, cross-platform, and used on Intel products to increase inference performance (lower latency and high throughput). Not only do the results show impressive performance benefits achieved through inference optimization techniques, the techniques themselves also have deep roots in theories of efficient information processing by intelligent systems. We therefore think that in the future they should not solely be seen as post-hoc strategies to improve models which have already been built but should instead consider them in the early steps of building a new model. Given the effectiveness of the techniques, considering inference optimization from the early stages of model building holds the promise of leading to a new wave of more efficient AI algorithms.

\clearpage
{\small
\bibliographystyle{ieeetr}
\bibliography{references}
}
\end{document}